\documentclass[10pt,twocolumn,letterpaper]{article}
\usepackage[utf8]{inputenc}
\usepackage{cvpr}
\usepackage{times}
\usepackage{epsfig}
\usepackage{graphicx}
\usepackage{amsmath}
\usepackage{amssymb}
\usepackage{numprint}
\usepackage{makecell}
\usepackage{balance}

\usepackage[breaklinks=true,bookmarks=false]{hyperref}

\cvprfinalcopy 

\def\cvprPaperID{48} 

\begin{document}

\title{CNNs Fusion for Building Detection in Aerial Images\\for the Building Detection Challenge}

\author{Remi Delassus\\
Qucit, 213 cours Victor Hugo, 33130 Begles, France\\
Univ. Bordeaux, Bordeaux INP, CNRS, LaBRI, UMR5800, F-33400 Talence, France\\
{\tt\small remi.delassus@qucit.com}
\and
Romain Giot\\
Univ. Bordeaux, Bordeaux INP, CNRS, LaBRI, UMR5800, F-33400 Talence, France\\
{\tt\small romain.giot@u-bordeaux.fr}
}

\maketitle
\thispagestyle{empty}

\begin{abstract}
This paper presents our contribution to the DeepGlobe Building Detection Challenge.
   We enhanced the SpaceNet Challenge winning solution by proposing a new fusion strategy based on a deep combiner using segmentation both results of different CNN and input data to segment. 
Segmentation results for all cities have been significantly improved (between 1\% improvement over the baseline for the smallest one to more than 7\% for the largest one). The separation of adjacent buildings 
should be the next enhancement made to the solution.
\end{abstract}

\section{Introduction}

The DeepGlobe Building Detection Challenge~\cite{demir2018deepglobe} follows the second round of the SpaceNet Challenge~\footnote{\url{https://community.topcoder.com/tc?module=MatchDetails&rd=16892}}. It poses the challenge of automatically
detecting buildings from satellite images

Top competitors of the SpaceNet challenge have submitted their code which is now publicly available~\footnote{\url{https://github.com/SpaceNetChallenge/BuildingDetectors\_Round2}}. We thus studied the winning solution that consists in an ensemble of three U-Net~\cite{ronneberger2015u} based models whose segmentation results are averaged, further called the ``baseline".

A model built by merging the prediction of multiple sub models, like this one, is called an ensemble model; it generally performs better than the best of the merged models~\cite{kittler_combining_1998}. The most common 
merging strategy for image segmentation is an unweighted average of the predictions.

We propose to replace the original unweighted average by another ensemble model that we a call a deep combiner. Its architecture is equivalent to the combined models (\emph{i.e}, U-Net based) and works with a MUL input as well as predicted segmentations. This combiner improves the baseline from 1\% in Vegas to 7.4\% in Khartoum.

This paper is organized as follows. 
Section~\ref{state_art} presents previous works on building detection and fusion techniques. We see here the existing building extraction methods, existing work on models' output fusion for segmentation, and SpaceNet's winning solution presentation.
Section~\ref{supers} presents our deep combiner.
Section~\ref{results} shows the results; the combiner enhance the segmentation and the detection of buildings.
Section~\ref{conclusion} concludes this work.

\section{Previous work}
\label{state_art}

We based our model on the winning solution of the SpaceNet challenge. it used an unweghted average based ensemble
of 3 U-Net models to segment buildings from the 8 MUL channels of the input Image as well as OpenStreetMap data. The contours of the building are then extractedd from the segmentation by grouping together into one building all connected pixels.

\subsection{Limitations of the baseline}

Several limits of the baseline appear. Firstly, the models are trained to predict a mask, which is not optimal.
When two buildings are too close to 
each other, the contours of the buildings cannot be extracted from the corresponding mask (\figurename~\ref{adjacent}) and the two buildings are considered as one.

Secondly, the post-processing is close to non existent. For example, one could use an instance segmentation 
model~\cite{bai2017deep} or a flooding algorithm~\cite{meyer1992color} to split adjacent buildings. Those adjacent buildings will be fused into one single building if a group of at least one pixel adjacent to both of them is classified as a group of building pixels. Another example of what can be done during the post processing phase is the smoothing of building edges. Because most of the time, buildings have regular shapes, we can use this knowledge to predict shapes closer to reality. CRF~\cite{krahenbuhl2011efficient} techniques have been used to smooth edges and enhance the resolution of segmentations. It could help sharpening polygons and separating them.   

Thirdly, more recent architectures such as DeepLab~\cite{chen2016deeplab} or PSPnet~\cite{zhao2016pyramid} could be 
used to try to enhance the segmentation score. They outperform the U-Net architecture in ImageNet~\cite{ILSVRC15} segmentation tasks. However, their strong performance comes from the ability to detect tiny objects as well as big ones, which may not be so useful in the building detection problem.

Finally, a simple unweighted average is often a good default choice, as it will reduce the variance of the result, but might 
not be the best fusion solution as it gives the same importance to all models regardless of their performance, and is subject
to errors when one of the merged models is overconfident, or when a large proportion of the merged models is erroneous. This critical proportion is easily reached when only three models are merged. 

This study focuses on this final limitation. We thus review the state of ensemble modeling applied to problems of image segmentation.

\subsection{Ensemble modeling for image segmentation}
 
Ju~\textit{et al}~\cite{ju2017relative} show that unweighted average (used by Marmanis~\textit{et al}~\cite{marmanis2016semantic} on multiple instances of the same model architecture and 
by Kamnitsas~\textit{et al}~\cite{kamnitsas2017ensembles} on complementary architectures) is a fusion method 
that performs as good as, if not better than, other known methods. It is favorable for reducing variance among 
predictions, but suffers from overconfident models. Majority voting 
(used by Dolz~\textit{et al}~\cite{dolz2017deep}) generally has a lower performance 
pixel wise, but can produce an uncertainty map, which can be used to focus the model training on uncertain 
areas~\cite{wang2017gated}, or to prevent uncertain pixel classifications~\cite{kampffmeyer2016semantic} 
which could help differentiate buildings close to each other and lead to worthier performances in building 
detection. 

Weighted average with learned weights is called by Ju~\textit{et al}
~\cite{ju2017relative} a super learner. It is a model that takes segmentation as an input and outputs a more advantageous one. 
The simplest one can be seen as $1*1*m$ convolution where $m$ is the number of 
models fusionned. In their study on the CIFAR10~\cite{krizhevsky2009learning} dataset, they show that it 
usually performs better than unweighted averages but does not investigate more complex super learners, such as 
the one studied here; super learners taking neighborhood into account, or deep super learners. They do 
not talk about non learned weights, such as weighs accordingly proportional to the models' 
performances~\cite{lahiri2016deep}, but we can reasonably assume that those weights would be reached by 
learn methods if they happened to be optimal, simply by setting them as initial weights.

The success of residual networks~\cite{he2016deep} shows that learning residual functions with reference to the layer inputs, instead of learning unreferenced functions can lead to better performance. Scaled up to the model instead of the layer, we study a model that will use the input image as well as the fused models' outputs to learn a better segmentation. This additional input puts our model outside of the super learner category. We will label as \textbf{combiners} those models dealing with combining the output of other models, with or without additional outputs.

\section{Proposed combiner}
\label{supers}

We propose a deep learning approach to the fusion of segmentation: a U-net based deep combiner that combines the segmentations output of the combined models as well as the MUL channels of the input image. 

\subsection{Architecture}

The Deep combiner
approach aims to enhance the segmentation by letting the deep neural network compute complex 
features such as edge detection, distance to building pixels etc. Moreover, this combiner uses not only the three predicted segmentations, but also the original image as input. This image can be used by the combiner model to determine the residuals of input models and fix their mistakes. We use the MUL channels of the original image as input.

We use the same U-Net architecture 
as input networks, with 29 hidden layers and $\numprint{7838273}$ parameters.
This architecture is usually used to segment images. Combining segmentations into one can be seen as 
performing a new segmentation on an image in which some channels correspond to some proposed segmentations. It thus makes
sense to use a segmentation architecture to perform the combination.

This combiner was not manually initialized to perform the unweighted average of the previous segmentation as it would jeopardize the learning process. Weigths are initialized with a uniform distribution between -0.05 and +0.05.
We name this combiner the \emph{U-net} approach.

\subsection{Training}
Training data is augmented, as we don't have enough images for some cities to train a deep learning model. For example, we train the model for Khartoum on 708 images and validate it on 304. The additional data comes from the rotation of the original image of 90, 180 and 270 degrees, as well as a symmetry along both axis. 

We train our combiner during 20 epochs with a batch size of 1. We selecte the best epoch based on the validation score. Optimal validation score is usually reached between epoch 5 and epoch 10. We evaluate the segmentation using the Jaccard coefficient~\cite{jaccard1901distribution} for our loss function. It is the closest
metric to the IoU score (used to judge if two polygons match) when it comes to image comparison.
Given a ground truth labeling $y^*$ and a predicted segmentation to be evaluated $\hat{y}$, the Jaccard coefficient
is defined as:

\begin{equation}
	J(y^*,\hat{y}) = \frac{y^* \cap \hat{y}}{y^* \cup \hat{y}}
\end{equation}

\noindent which for images translates into: 

\begin{equation}
	J(y^*,\hat{y}) = \frac{\sum{(y^* * \hat{y})}}{\sum{(y^* + \hat{y})} - \sum{(y^* * \hat{y})}}
\end{equation}

With the assumption that $J(0, 0) = 1$. 
As the Jaccard coefficient produces a value between 0 and 1, it is straightforward to transform it into a loss function: 

\begin{equation}
J_l(y^*,\hat{y}) = 1 - J(y^*,\hat{y})
\end{equation}

Experiments have been run on a Linux computer with the following characteristics: Intel(R) Core(TM) i7-6950X 
CPU @ 3.00GHz, three Titan X Pascal GPUs (used to train multiple model in parallel, with one model per GPU) and 120Gbits of RAM, using Keras~\cite{chollet2015keras} 1.2.2 with Tensorflow~\cite{abadi2016tensorflow} 1.4.0.

\section{Results}
\label{results}

\subsection{F-score improvement}

Table~\ref{F_scores} reports the maximum value of F-score obtained on each city dataset with the baseline and the combiner.  The gain is defined as $Gain= (New\_Score - Baseline) / Baseline$. There has only been a slight improvement on Vegas because the baseline is already performing extremely well. However a large improvement on Khartoum shows that our solution can address more cases than a simple unweighted average.

\begin{table}[tb]
	\centering
	\caption{\label{F_scores} F-score, of the baseline and U-Net combiners over the validation set (including U-Net gain over baseline) compared to U-Net combiner final score for this challenge.
	}
	\begin{tabular}{|l|p{0.07\textwidth}|p{0.07\textwidth}|c|}
		\hline
		\textbf{City} &
		\textbf{Baseline} &
		\textbf{U-Net} &
        \textbf{U-Net final score}
		\\
		\Xhline{2\arrayrulewidth}
		\hline
		Vegas& 0.8559 \newline - & 0.8639 \newline +0.93\% & 0.8057 \\
		\hline
		Paris& 0.6805 \newline -& 0.7080 \newline  +4.04\% & 0.6787 \\
		\hline
		Shanghai& 0.5627 \newline - & 0.5794 \newline  +2.97\% & 0.5661 \\
		\hline
		Khartoum& 0.5855 \newline - &  0.6290 \newline  +7.43\% & 0.6387 \\
		\hline
	\end{tabular}
\end{table}

\subsection{Visualization of Detected Buildings}

\begin{figure}
\centering 
	\includegraphics[width=0.8\linewidth]{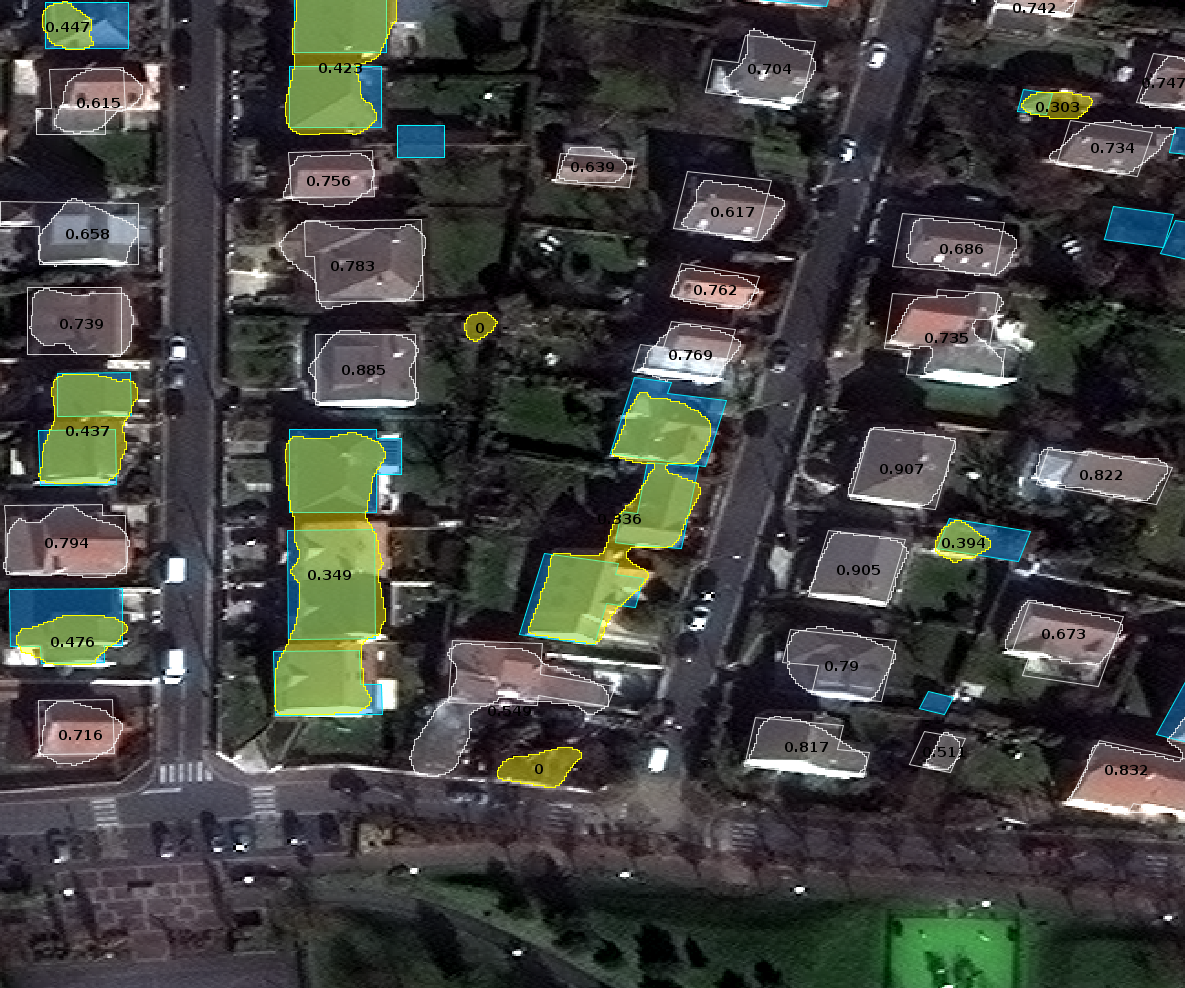}
	\includegraphics[width=0.8\linewidth]{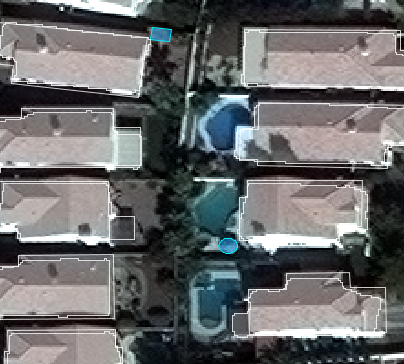}
	\caption{\label{result_image}Example of errors.
In the first image adjacent buildings are fused because they were too close from each other. In the second image only the big buildings have been detected while the little roofs have not.
}
\end{figure}

\begin{figure}
\centering 
	\includegraphics[width=0.8\linewidth]{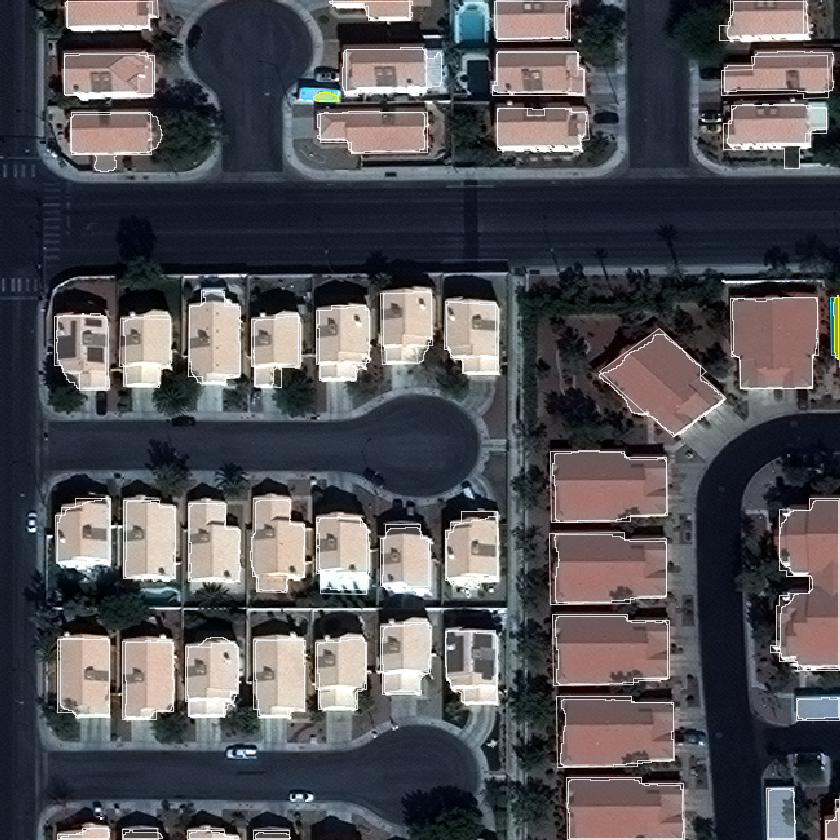}
	\includegraphics[width=0.8\linewidth]{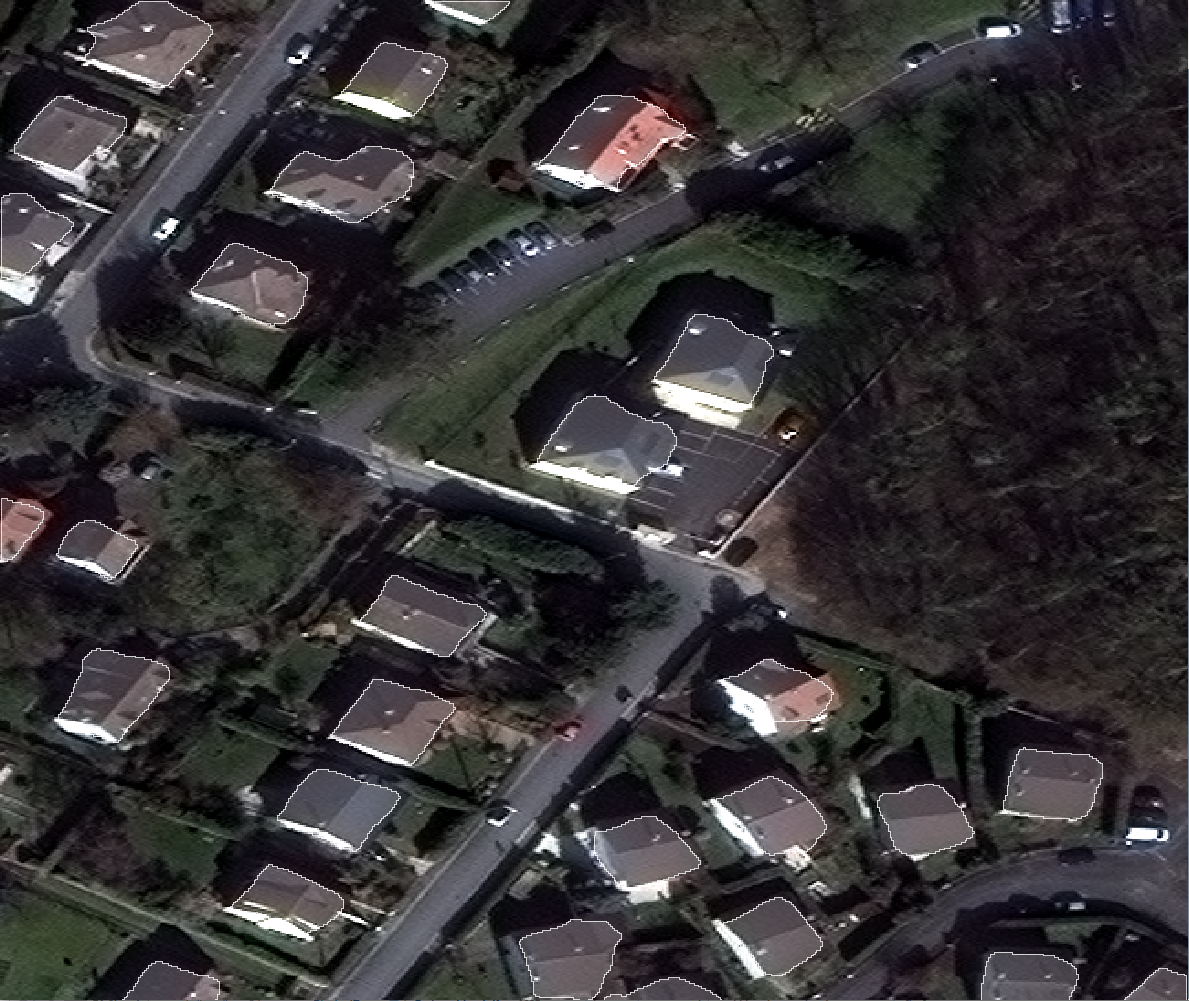}
	\caption{\label{good_image}Example of high building density images. We can see that a large majority of the buildings are correctly detected. The few errors come from the smaller buildings, or from small parts of buildings cut at the edges of the image, as well as adjacent buildings being merged into one.
}
\end{figure}

Images in Figures~\ref{result_image} and ~\ref{good_image} have been produced with the SpaceNet Challenge Building Detector Visualizer~\footnote{\url{https://github.com/SpaceNetChallenge/BuildingDetectorVisualizer}}.
Ground truth and predicted polygons are overlaid over the RGB images.
Matching ground truth and predicted polygons are white. Yellow polygons are false positives (predicted polygon with no matching ground truth). Blue polygons are false negatives (ground truth polygon with no matching prediction).

A vast majority of buildings are properly segmented. Only buidings with a surface area significantly smaller than surrounding buildings are ignored by the combiner. The major difficulty is to separate buildings too close from each other.

\subsection{Discussion}
\label{discussion}

\begin{figure}
\centering
	\includegraphics[width=0.85\linewidth]{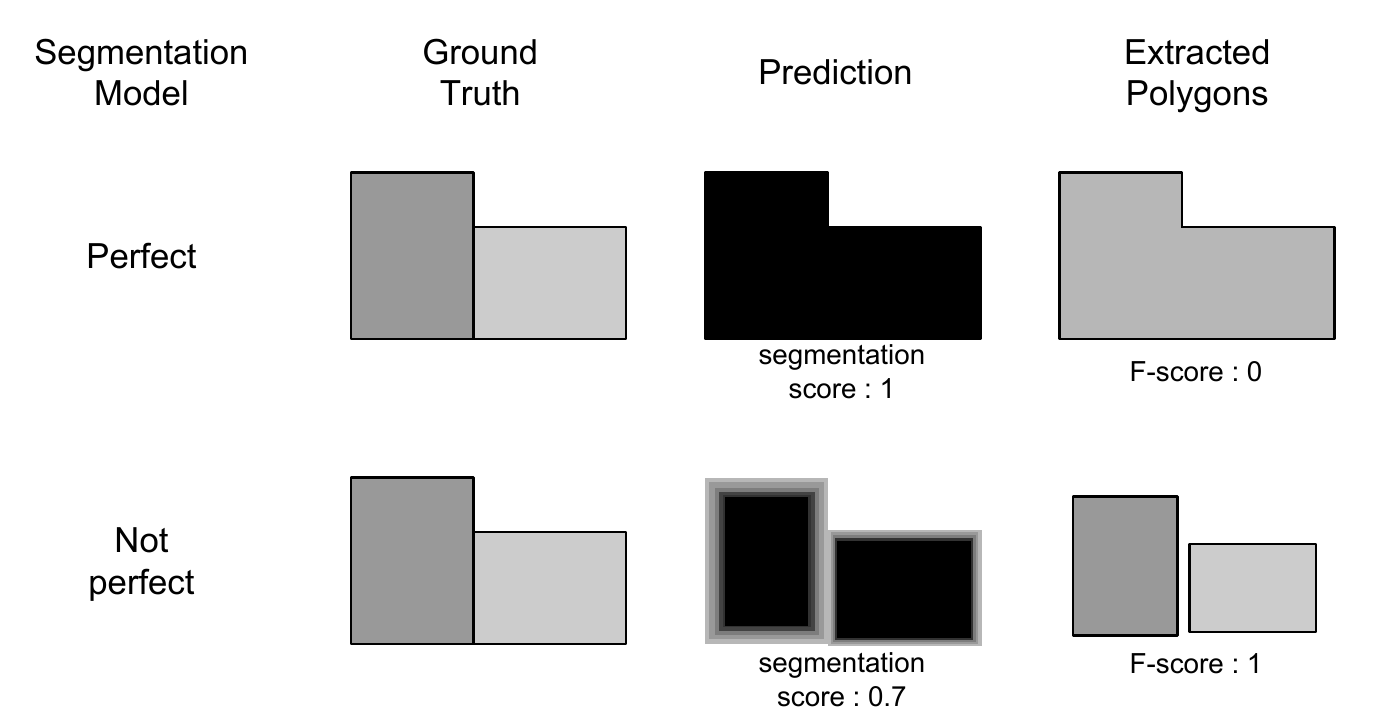}
	\caption{\label{better_worse}\label{adjacent}Adjacent buildings problem. When two buildings are too close to each other, a perfect model will produce only one building if the chosen label is a mask.}
\end{figure}

We have been able to highlight two main sources of errors. The first one is the adjacent buildings problem (Figure~\ref{better_worse}). When 
we reduce the model uncertainty around buildings, adjacent buildings are considered as one unique building instead of two different ones.
The second one comes from numerous errors in labeled data. Those errors can prevent the models to learn useful features or can punish them when they do. Missing a lot of small buildings will have a huge impact on the F-score, but smaller buildings are not consistenly labeled, especially in Shanghai.

To solve the adjacent buildings problem,
Marmanis~\textit{et al}~\cite{marmanis2016classification} introduce the border class, such that the model 
classifies each pixel in three classes: building, building border, and background. It is then easier to split 
adjacent buildings. Yuan~\textit{et al}~\cite{yuan2017learning} use a distance field as label. Each pixel value is its distance to the closest border (negative when inside a building). It is then possible 
to configure the post-processing and set a predicted distance (0 in the mask) as the border. But it requires 
a large receptive field for this dataset.

We used the same architecture for the combiner as the one used for the combined models. More architectures 
should be tested to determine if regular convnets are the best combiners. 
Maggiori~\textit{et al}~\cite{maggiori2017recurrent} used a recurrent neural network to improve their 
segmentation, which is another solution. Bai~\textit{et al}~\cite{bai2017deep} used a new architecture 
of convnets to perform an instance segmentation. Their model takes the input image and the result of a 
segmentation as an input, just like our U-net.

The deep combiner was only tested with the combined learned output segmentation and the MUL input. 
The OSM layers and multi band inputs were not used.
Of the three combined learners, the best ones don't use RGB, but multiband image and OSM layers. 
A deep combiner using those inputs is likely to have better performances too and will thus be implemented
once the adjacent buildings' problem will be solved.

\section{Conclusion}
\label{conclusion}
This paper addresses the problem of segmenting buildings from aerial images.
It proposes a fusion solution for an ensemble of U-net models used to segment aerial images and
extract buildings' contours from the segmentation.
We used SpaceNet's challenge winning solution as a baseline to evaluate the benefits of combiners 
over unweighted averages for model fusion. This baseline corresponds to an ensemble of three models with the same 
architecture but different inputs. For two of the three studied images set (Vegas and Khartoum), a better segmentation led to a better F-
score, and combiners were thus performing superiorly than the existing unweighted averages. For the Paris dataset, the 
enhancement of the segmentation led to a worsened F-score, showing that the separation of adjacent buildings 
should be the next part of the solution to enhance. 

Future work will focus on adjacent buildings' separation, thanks to an advantageous label such as the addition of a 
\textit{building borders} class, or through post processing of the segmentation, as proposed by 
Bai~\textit{et al}~\cite{bai2017deep}.

{\small
\balance
\bibliographystyle{ieee}
\bibliography{egbib}

\begin{thebibliography}{10}\itemsep=-1pt

\bibitem{abadi2016tensorflow}
M.~Abadi, A.~Agarwal, P.~Barham, E.~Brevdo, Z.~Chen, C.~Citro, G.~S. Corrado,
  A.~Davis, J.~Dean, M.~Devin, et~al.
\newblock Tensorflow: Large-scale machine learning on heterogeneous distributed
  systems.
\newblock {\em arXiv preprint arXiv:1603.04467}, 2016.

\bibitem{bai2017deep}
M.~Bai and R.~Urtasun.
\newblock Deep watershed transform for instance segmentation.
\newblock In {\em CVPR, 2017}, pages 2858--2866. IEEE, 2017.

\bibitem{chen2016deeplab}
L.-C. Chen, G.~Papandreou, I.~Kokkinos, K.~Murphy, and A.~L. Yuille.
\newblock Deeplab: Semantic image segmentation with deep convolutional nets,
  atrous convolution, and fully connected crfs.
\newblock {\em arXiv preprint arXiv:1606.00915}, 2016.

\bibitem{chollet2015keras}
F.~Chollet et~al.
\newblock Keras.
\newblock \url{https://github.com/fchollet/keras}, 2015.

\bibitem{demir2018deepglobe}
I.~Demir, K.~Koperski, D.~Lindenbaum, G.~Pang, J.~Huang, S.~Basu, F.~Hughes,
  D.~Tuia, and R.~Raskar.
\newblock Deepglobe 2018: A challenge to parse the earth through satellite
  images.
\newblock {\em arXiv preprint arXiv:1805.06561}, 2018.

\bibitem{dolz2017deep}
J.~Dolz, C.~Desrosiers, L.~Wang, J.~Yuan, D.~Shen, and I.~B. Ayed.
\newblock Deep cnn ensembles and suggestive annotations for infant brain mri
  segmentation.
\newblock {\em arXiv preprint arXiv:1712.05319}, 2017.

\bibitem{he2016deep}
K.~He, X.~Zhang, S.~Ren, and J.~Sun.
\newblock Deep residual learning for image recognition.
\newblock In {\em Proceedings of the IEEE conference on computer vision and
  pattern recognition}, pages 770--778, 2016.

\bibitem{jaccard1901distribution}
P.~Jaccard.
\newblock Distribution de la flore alpine dans le bassin des dranses et dans
  quelques r{\'e}gions voisines.
\newblock {\em Bull. Soc. Vaud. Sci. Nat.}, 37:241--272, 1901.

\bibitem{ju2017relative}
C.~Ju, A.~Bibaut, and M.~J. van~der Laan.
\newblock The relative performance of ensemble methods with deep convolutional
  neural networks for image classification.
\newblock {\em arXiv preprint arXiv:1704.01664}, 2017.

\bibitem{kamnitsas2017ensembles}
K.~Kamnitsas, W.~Bai, E.~Ferrante, S.~McDonagh, M.~Sinclair, N.~Pawlowski,
  M.~Rajchl, M.~Lee, B.~Kainz, D.~Rueckert, et~al.
\newblock Ensembles of multiple models and architectures for robust brain
  tumour segmentation.
\newblock {\em arXiv preprint arXiv:1711.01468}, 2017.

\bibitem{kampffmeyer2016semantic}
M.~Kampffmeyer, A.-B. Salberg, and R.~Jenssen.
\newblock Semantic segmentation of small objects and modeling of uncertainty in
  urban remote sensing images using deep convolutional neural networks.
\newblock In {\em Proceedings CVPR Workshops}, pages 1--9, 2016.

\bibitem{kittler_combining_1998}
J.~Kittler, M.~Hatef, R.~P.~W. Duin, and J.~Matas.
\newblock On combining classifiers.
\newblock {\em IEEE PAMI}, 20(3):226--239, Mar. 1998.

\bibitem{krahenbuhl2011efficient}
P.~Kr{\"a}henb{\"u}hl and V.~Koltun.
\newblock Efficient inference in fully connected crfs with gaussian edge
  potentials.
\newblock In {\em Advances in neural information processing systems}, pages
  109--117, 2011.

\bibitem{krizhevsky2009learning}
A.~Krizhevsky and G.~Hinton.
\newblock Learning multiple layers of features from tiny images.
\newblock 2009.

\bibitem{lahiri2016deep}
A.~Lahiri, A.~G. Roy, D.~Sheet, and P.~K. Biswas.
\newblock Deep neural ensemble for retinal vessel segmentation in fundus images
  towards achieving label-free angiography.
\newblock In {\em Engineering in Medicine and Biology Society (EMBC), 2016 IEEE
  38th Annual International Conference of the}, pages 1340--1343. IEEE, 2016.

\bibitem{maggiori2017recurrent}
E.~Maggiori, G.~Charpiat, Y.~Tarabalka, and P.~Alliez.
\newblock Recurrent neural networks to correct satellite image classification
  maps.
\newblock {\em IEEE TGRS}, 2017.

\bibitem{marmanis2016classification}
D.~Marmanis, K.~Schindler, J.~D. Wegner, S.~Galliani, M.~Datcu, and U.~Stilla.
\newblock Classification with an edge: improving semantic image segmentation
  with boundary detection.
\newblock {\em arXiv preprint arXiv:1612.01337}, 2016.

\bibitem{marmanis2016semantic}
D.~Marmanis, J.~D. Wegner, S.~Galliani, K.~Schindler, M.~Datcu, and U.~Stilla.
\newblock Semantic segmentation of aerial images with an ensemble of cnss.
\newblock {\em ISPRS Annals of the Photogrammetry, 2016}, 3:473--480, 2016.

\bibitem{meyer1992color}
F.~Meyer.
\newblock Color image segmentation.
\newblock In {\em Image Processing and its Applications, 1992., International
  Conference on}, pages 303--306. IET, 1992.

\bibitem{ronneberger2015u}
O.~Ronneberger, P.~Fischer, and T.~Brox.
\newblock U-net: Convolutional networks for biomedical image segmentation.
\newblock In {\em MICCAI}, pages 234--241. Springer, 2015.

\bibitem{ILSVRC15}
O.~Russakovsky, J.~Deng, H.~Su, J.~Krause, S.~Satheesh, S.~Ma, Z.~Huang,
  A.~Karpathy, A.~Khosla, M.~Bernstein, A.~C. Berg, and L.~Fei-Fei.
\newblock {ImageNet Large Scale Visual Recognition Challenge}.
\newblock {\em International Journal of Computer Vision (IJCV)},
  115(3):211--252, 2015.

\bibitem{wang2017gated}
H.~Wang, Y.~Wang, Q.~Zhang, S.~Xiang, and C.~Pan.
\newblock Gated convolutional neural network for semantic segmentation in
  high-resolution images.
\newblock {\em Remote Sensing}, 9(5):446, 2017.

\bibitem{yuan2017learning}
J.~Yuan.
\newblock Learning building extraction in aerial scenes with convolutional
  networks.
\newblock {\em IEEE transactions on pattern analysis and machine intelligence},
  2017.

\bibitem{zhao2016pyramid}
H.~Zhao, J.~Shi, X.~Qi, X.~Wang, and J.~Jia.
\newblock Pyramid scene parsing network.
\newblock {\em arXiv preprint arXiv:1612.01105}, 2016.

\end{thebibliography}
}

\end{document}